\documentclass{article}

\usepackage[numbers]{natbib}
\usepackage[final]{nips_2017}

\usepackage[utf8]{inputenc} % allow utf-8 input
\usepackage[T1]{fontenc}    % use 8-bit T1 fonts
\usepackage{hyperref}       % hyperlinks
\usepackage{url}            % simple URL typesetting
\usepackage{booktabs}       % professional-quality tables
\usepackage{amsfonts}       % blackboard math symbols
\usepackage{nicefrac}       % compact symbols for 1/2, etc.
\usepackage{microtype}      % microtypography
\usepackage{amsmath,graphicx}
\usepackage{amsfonts}
\usepackage{float}

\title{End-to-End Optimization of Task-Oriented Dialogue Model with Deep Reinforcement Learning}

\author{
	Bing Liu$^1$\thanks{Work done while the author was an intern at Google Research.}, Gokhan T\"{u}r$^2$, Dilek Hakkani-T\"{u}r$^2$, Pararth Shah$^2$, Larry Heck$^2$ \\
    $^1$Carnegie Mellon University, Pittsburgh, PA 15213 \\
    $^2$Google Research, Mountain View, CA 94043 \\
    \texttt liubing@cmu.edu, \{gokhant,dilekh,pararth,larryheck\}@google.com
}

\begin{document}

\maketitle

\begin{abstract}
    In this paper, we present a neural network based task-oriented dialogue system that can be optimized end-to-end with deep reinforcement learning (RL). The system is able to track dialogue state, interface with knowledge bases, and incorporate query results into agent's responses to successfully complete task-oriented dialogues. Dialogue policy learning is conducted with a hybrid supervised and deep RL methods. We first train the dialogue agent in a supervised manner by learning directly from task-oriented dialogue corpora, and further optimize it with deep RL during its interaction with users. In the experiments on two different dialogue task domains, our model demonstrates robust performance in tracking dialogue state and producing reasonable system responses. We show that deep RL based optimization leads to significant improvement on task success rate and reduction in dialogue length comparing to supervised training model. We further show benefits of training task-oriented dialogue model end-to-end comparing to component-wise optimization with experiment results on dialogue simulations and human evaluations. 
\end{abstract}

\section{Introduction}
    Task-oriented dialogue, different from chit-chat type of conversation, requires the system to produce responses by accessing information from knowledge bases and planning over multiple dialogue turns. Conventional task-oriented dialogue systems have a complex pipeline \cite{raux2005let,young2013pomdp} consisting of independently developed and modularly connected components for natural language understanding (NLU) \cite{mesnil2015using,Liu2016}, dialogue state tracking (DST) \cite{henderson2014word,mrkvsic2016neural}, and dialogue policy \cite{gasic2014gaussian,su2016line}. A limitation with such pipelined design is that errors made in upper stream modules may propagate to downstream components, making it hard to identify and track the source of errors. Moreover, each component in the pipeline is ideally re-trained as preceding components are updated, so that we have inputs similar to the training examples at run-time. This domino effect causes several issues in practice.
    
    To ameliorate these limitations with the conventional pipeline dialogue systems, recent efforts have been made in designing neural network based end-to-end learning solutions. Such end-to-end systems aim to optimize directly towards final system objectives (e.g. response generation, task success rate) instead of performing component-wise optimization. Many of the recently proposed end-to-end models are trained in supervised manner \cite{wenN2N16,bordes2017,eric2017copy,Liu2017} by learning from human-human or human-machine dialogue corpora. Deep reinforcement learning (RL) based systems \cite{li2017end,liu2017iterative,williams2017hybrid,dhingra2017towards} that learns by interacting with human user or user simulator have also been studied in the literature. Comparing to supervised training models, systems trained with deep RL showed improved task success rate and model robustness towards diverse dialogue scenarios. 
    
    In this work, we present a neural network based task-oriented dialogue system that can be optimized end-to-end with deep RL. The system is built with neural network components for natural language encoding, dialogue state tracking, and dialogue policy learning. Each system component takes in underlying component's outputs in a continuous from which is fully differentiable with respect to the system optimization target, and thus the entire system can be trained end-to-end. In the experiments on a movie booking domain, we show that our system trained with deep RL leads to significant improvement on dialogue task success rate comparing to supervised training systems. We further illustrate the benefit of performing end-to-end optimization comparing to only updating the policy network during online policy learning as in many previous work \cite{gasic2014gaussian,su2016line}.
    
\section{Related Work}
	Traditional task-oriented dialogue systems typically require a large number of handcrafted features, making it hard to extend a system to new application domains. Recent approaches to task-oriented dialogue treat the task as a partially observable Markov Decision Process (POMDP) \cite{young2013pomdp} and use RL for online policy optimization by interacting with users \cite{gavsic2013line}. The dialogue state and action space have to be carefully designed in order to make the reinforcement policy learning tractable \cite{young2013pomdp}.
    
    With the success of end-to-end trainable neural network models in modeling non-task-oriented chit-chat dialogues \cite{serban2015building,li2016deep}, efforts have been made in carrying over the good performance of end-to-end models to task-oriented dialogues. Bordes and Weston \cite{bordes2017} proposed modeling task-oriented dialogues with a machine reading approach using end-to-end memory networks. Their model removes the dialogue state tracking module and selects the final system response directly from candidate responses. Comparing to this approach, our model explicitly tracks user's goal in dialogue state over the sequence of turns, as robust dialogue state tracking has been shown \cite{jurvcivcek2012reinforcement,dhingra2017towards} to be useful for interfacing with a knowledge base (KB) and improving task success rate. Wen et al.~\cite{wenN2N16} proposed an end-to-end trainable neural network model with modularly connected system components. This system is trained in a supervised manner, and thus may not be robust enough to handle diverse dialogue situations due to the limited varieties in the training dialogue samples. Our system is trained by a combination of SL and deep RL methods, as it is shown that RL training may effectively improve the system robustness and dialogue success rate \cite{li2017end,williams2017hybrid}. Dhingra et al.~\cite{dhingra2017towards} proposed an end-to-end RL dialogue agent for information access. Their model focuses on bringing differentiability to the KB query operation by introducing a "soft" retrieval process in selecting the KB entries. Such soft-KB lookup may be prone to information updates in the KB, which is common in real world information systems. In our model, we use symbolic query and leave the selection of KB entities to external services (e.g. a recommender system), as entity ranking in real world systems can be made with much richer feature sets (e.g. user profiles, location and time context, etc.). Quality of the generated query is directly related to the performance of our dialog state tracking module, which can be optimized during user interactions in the proposed end-to-end reinforcement learning model.

\section{Proposed Method}
\subsection{System Architecture}
\label{sec:method}
\begin{figure*}[t]
    \centering
    \includegraphics[width=220pt]{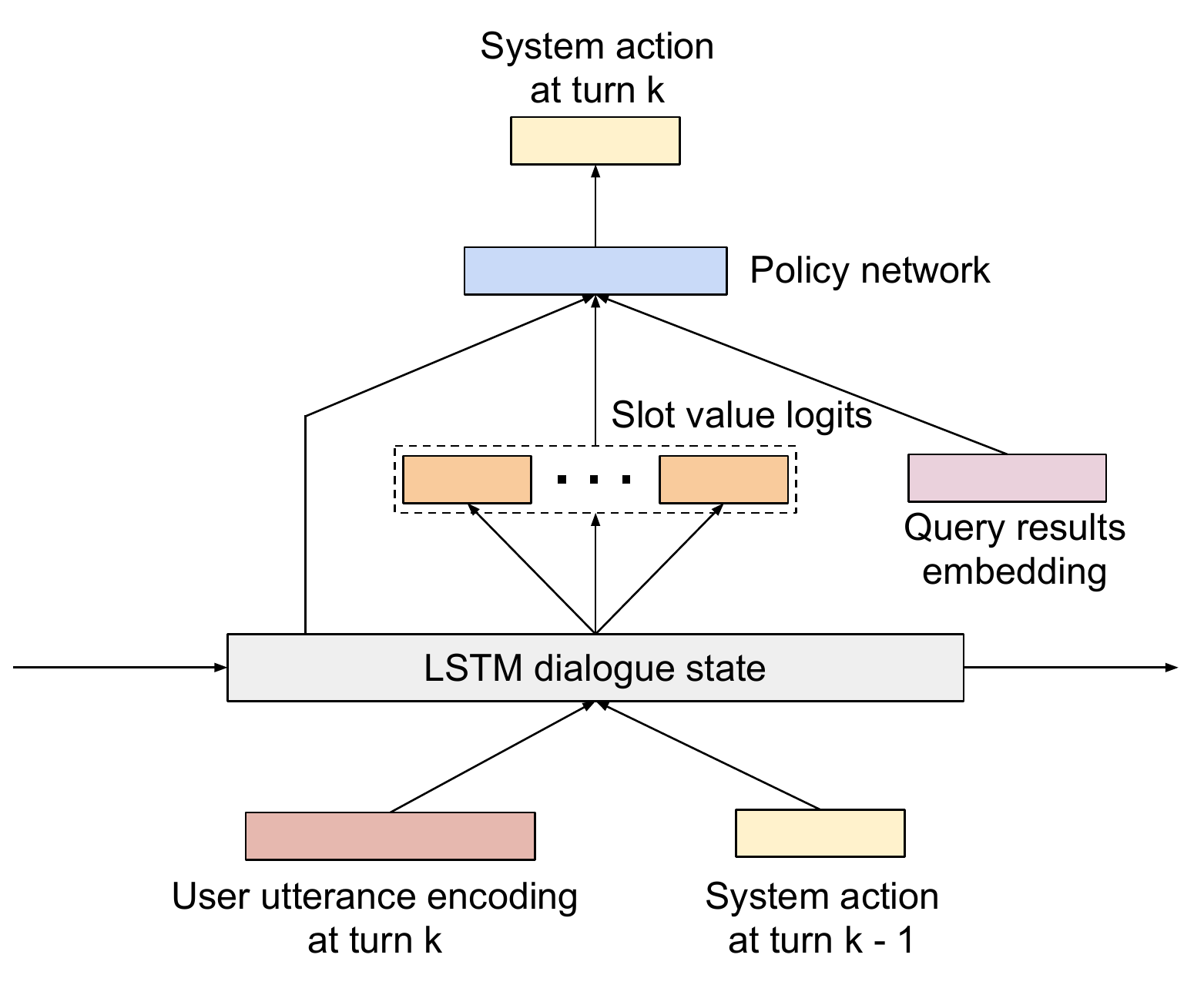}
    \vspace*{-2ex}
    \caption{{ Proposed end-to-end task-oriented dialogue model architecture. }}
    \label{fig:e2e_dialogue_nn}
    \vspace*{-2ex}
\end{figure*}

	Figure \ref{fig:e2e_dialogue_nn} shows the overall system architecture of the proposed end-to-end task-oriented dialogue model. A continuous form dialogue state over a sequence of turns is maintained in the state $s_k$ of a dialogue-level LSTM. At each dialogue turn $k$, this dialogue-level LSTM takes in the encoding of the user utterance $U_k$ and the encoding of the previous system action $A_{k-1}$, and produces a probability distribution $P(l^{m}_k)$ over candidate values for each of the tracked goal slots:
        \begin{align}
            & s_k = \operatorname{LSTM}(s_{k-1}, \hspace{1mm} [U_k, \hspace{1mm} A_{k-1}]) \\
            & P(l^{m}_k \hspace{1mm} | \hspace{1mm} \mathbf{U}_{\le k}, \hspace{1mm} \mathbf{A}_{< k}) = \operatorname{SlotDist}_{m}(s_k)
        \end{align}
    where $\operatorname{SlotDist}_{m}$ is a single hidden layer MLP with $\operatorname{softmax}$ activation function over slot type $m \in M$. In encoding natural language user utterance to a continuous vector $U_k$, we use a bidirectional LSTM (i.e. an utterance-level LSTM) reader by concatenating the last forward and backward LSTM states. 

    Based on slot-value pair outputs from dialogue state tracking, a query command is formulated by filling a query template with candidate values that have the highest probability for each tracked goal slot. Alternatively, an n-best list of queries can be generated with the most probable candidate values. The query is sent to a KB to retrieve user requested information. Finally, a system action is emitted in response to the user's input based on the current dialogue state and the information retrieved from the knowledge base: 
    \begin{align}
        P(a_{k} \hspace{1mm} | \hspace{1mm} U_{\le k}, \hspace{1mm} A_{< k}, \hspace{1mm} E_{\le k}) = \operatorname{PolicyNet}(s_{k}, v_{k}, E_{k})
    \end{align}
    where $v_{k}$ represents the concatenated log probabilities of candidate values for each goal slot. $E_{k}$ is the encoding of the retrieved result from the knowledge base (e.g. item availability and number of matched items). $\operatorname{PolicyNet}$ is an MLP with $\operatorname{softmax}$ activation function over all system actions. The emitted system action is then translated to a system response in natural language format by combining the state tracking outputs and the query results. We use a template based natural language generator (NLG) in this work.

\subsection{Model Training}
    We first train the system in a supervised manner using task-oriented dialogue corpora. Based on system inputs with past user utterances, system actions, and KB results, the model tracks the user's goal slot values and predict the next system action. We optimize the model to minimize the linear interpolation of cross-entropy losses for dialogue state tracking and system action prediction: 
	\begin{equation}
            \begin{split}
            \min_{\theta} \sum_{k=1}^{K} -\Big[ \sum_{m=1}^{M} &\lambda _{l^{m}} \log P({l^{m}_k}^{*} | \mathbf{U}_{\le k}, \mathbf{A}_{< k}, \mathbf{E}_{< k}; \theta) \\
            + &\lambda _a \log P(a_k^{*} | \mathbf{U}_{\le k}, \mathbf{A}_{< k}, \mathbf{E}_{\le k}; \theta) \hspace{1mm} \Big] \\
            \end{split}
        \end{equation}
    where $\lambda$s are the linear interpolation weights for the cost of each system output. ${l^{m}_k}^{*}$ and $a_k^{*}$ are the ground truth labels for goal slots and system action the $k$th turn.

    After the supervised training stage, we further optimize the system with RL by letting the agent to interact with users and collecting user feedback. 
    We apply REINFORCE algorithm \cite{williams1992simple} in optimizing the network parameters. We use softmax policy during RL training to encourage the agent to explore the dialogue action space. Feedback is only collected at the end of a dialogue. A positive reward is assigned for success tasks, and a zero reward is assigned for failure tasks. A small step penalty is applied to each dialogue turn to encourage the agent to complete the task in fewer steps. We use policy gradient method for dialogue policy learning. With likelihood ratio gradient estimator, the gradient of the objective function $J_k(\theta)$ can be derived as:
        \begin{equation}
            \begin{split}
            \nabla  _{\theta} J_k(\theta) = \nabla _{\theta} \mathbb E_{\theta}\left[ R_k \right]
            % &= \sum_{a_{k}}  \pi _{\theta}(a_{k} | s_{k}) \nabla _{\theta} \log \pi _{\theta}(a_{k} | s_{k}) R_{k} \\
            = \mathbb E_{\theta_a}\left[ \nabla _{\theta} \log \pi _{\theta}(a_{k} | s_{k}) R_{k} \right]
            \end{split}
        \end{equation}
    This last expression above gives us an unbiased gradient estimator. We sample the agent action based on the currently learned policy at each dialogue turn and compute the gradient. 
    
\section{Experiments}
\subsection{Datasets}
    We evaluate the proposed method on DSTC2 \cite{henderson2014second} dataset in restaurant search domain and an internally collected dialogue corpus in movie booking domain. The movie booking corpus is generated with rule based dialogue agent and user simulator. The same user simulator is used to interact with our end-to-end learning agent during RL training. We use an extended set of NLG templates during model testing to evaluate the end-to-end model's generalization capability in handling diverse natural language inputs.
    
\subsection{Training Settings}
    We set state size of the dialogue-level and utterance-level LSTM as 200 and 150 respectively. Hidden layer size of the policy network is set as 100. We used randomly initialized word embedding of size 300. Adam optimization method \cite{kingma2014adam} with initial learning rate of 1e-3 is used for mini-batch training. Dropout rate of 0.5 is applied during training to prevent the model from over-fitting. 
    
    In dialogue simulation, we take a task-oriented dialogue as successful if the goal slot values estimated by the state tracker fully match to the user's true goal values, and the system is able to offer an entity which is finally accepted by the user. Maximum allowed number of dialogue turn is set as 15. A positive reward of +15.0 is given to the agent at the end of a success dialogue, and a zero reward is given in a failure case. We apply a step penalty of -1.0 for each turn to encourage shorter dialogue in completing the task.  
    
\subsection{Results and Analysis}
	Table \ref{tab:table_dstc2_sl} and Table \ref{tab:table_movie_corpus_sl} show the supervised training model performance on DSTC2 and the movie booking dialogue dataset. The model is evaluated on dialogue state tracking accuracy. On DSTC2 dataset, our end-to-end model achieves near-state-of-the-art state tracking performance comparing to the recent published results using RNN \cite{henderson2014robust} and NBT \cite{mrkvsic2016neural}. On the movie booking dataset, our model also achieves promising performance on individual slot tracking and joint slot tracking accuracy.

    \begin{table}[th]
      \caption{Belief tracking results on DSTC2 corpus (with ASR hypothesis as input)}
      \label{tab:table_dstc2_sl}
      \centering
      \begin{tabular}{l c c c c}
        \hline  
        \textbf{Model} & \textbf{Area}  & \textbf{Food}  & \textbf{Price}  & \textbf{Joint} \\
        \hline
        RNN \cite{henderson2014robust}    & 92 & 86 & 86 & 69  \\
        % RNN + sem. dict \cite{henderson2014robust}    & 92 & 86 & 92 & 71  \\
        NBT \cite{mrkvsic2016neural}    & 90 & 84 & 94 & 72  \\
        Our end-to-end model          & 90 & 84 & 92 & 72  \\
        \hline
      \end{tabular}     
    \end{table}

    \begin{table}[th]
      \caption{Belief tracking results on movie booking dataset}
      \label{tab:table_movie_corpus_sl}
      \centering
      \begin{tabular}{l c c c c c c}
        \hline  
        \textbf{Model} & \textbf{Num\_ticket} & \textbf{Movie} & \textbf{Theater} & \textbf{Date} & \textbf{Time} & \textbf{Joint} \\
        \hline  
        Our end-to-end model           & 98.22 & 91.86  & 97.33  & 99.31   & 97.71   & 84.57 \\
        \hline
        \vspace*{-1ex}
      \end{tabular}     
    \end{table}

\begin{figure}[t]
          \centering
          \includegraphics[width=\linewidth]{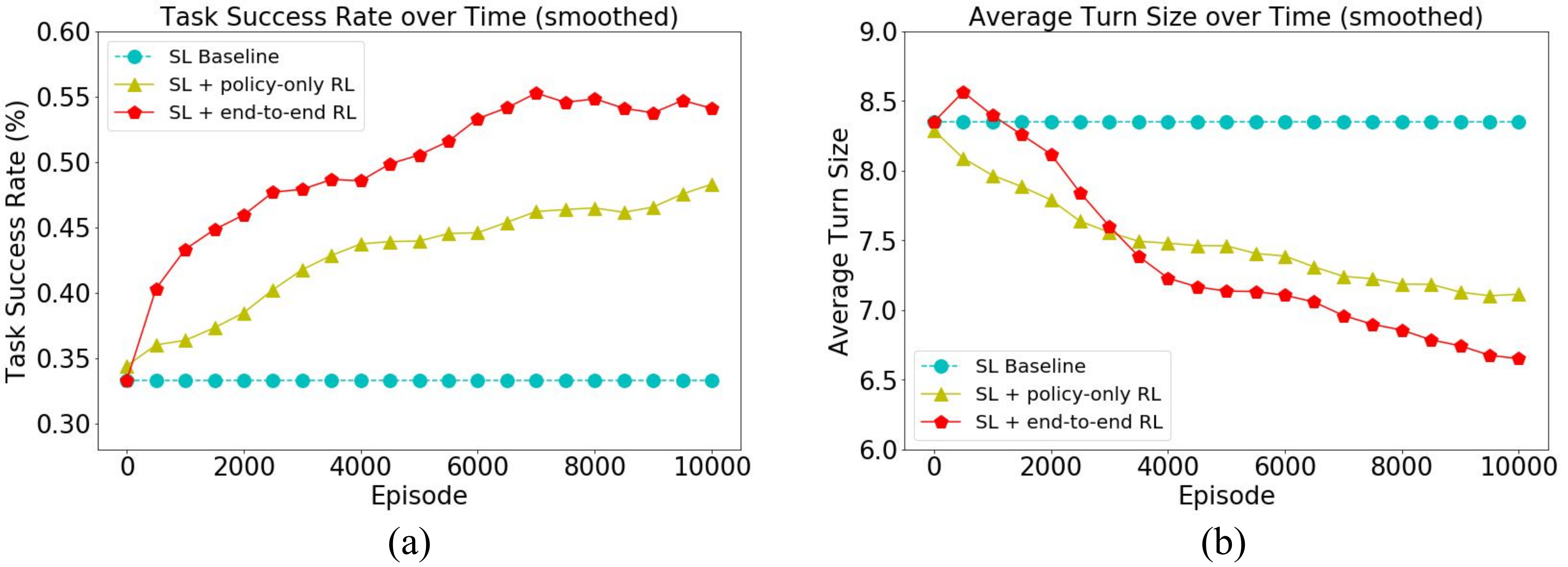}
          \vspace*{-4ex}
          \caption{RL curves on (a) dialogue task success rate and (b) average dialogue turn size.}
          \label{fig:e2e_training_curves}
        \end{figure}

	Figure \ref{fig:e2e_training_curves} shows the RL curves of the proposed model on dialogue task success rate and average dialogue turn size. Evaluation is based on dialogue simulations between our proposed end-to-end dialogue agent and the rule based user simulator. This is different from the evaluations based on fixed dialogue corpora as in Table \ref{tab:table_dstc2_sl} and \ref{tab:table_movie_corpus_sl}. The policy gradient based RL training is performed on top of the supervised training model. We compare models with two RL training settings, the end-to-end training and the policy-only training, to the baseline supervised learning (SL) model. 
     
    As shown in Figure \ref{fig:e2e_training_curves}(a), the SL model performs poorly during user interaction, indicating the limited generalization capability of the SL model to unseen dialogue state. Any mistake made by the agent during user interaction may lead to deviation of the dialogue from the training dialogue trajectories and states. The SL agent does not know how to recover from an unknown state, which leads to final task failure. RL model training, under both end-to-end learning and policy-only learning settings, continuously improves the task success rate with the growing number of user interactions. We see clear advantage of performing end-to-end model update in achieving higher dialogue task success rate comparing to only updating the policy network during interactive learning. 
     
    Figure \ref{fig:e2e_training_curves}(b) shows the learning curves for the average number of turns in successful dialogues. We observe decreasing number of dialogue turns along the growing number of interactive learning episodes. This shows that the dialogue agent learns better strategies to successfully complete the task in fewer numbers of turns. Similar to the results for task success rate, the end-to-end training model outperforms the model with policy-only optimization during RL training, achieving lower average number of dialogue turns in successfully completing a task. 

\subsection{Human Evaluations}
    We further evaluate our proposed method with human judges recruited via Amazon Mechanical Turk. Each judge is asked to read a dialogue between our model and the user simulator and rate each system turn on a scale of 1 (frustrating) to 5 (optimal way to help the user). Each turn is rated by 3 different judges. We rate the three models with 100 dialogues each: (i) the SL model, (ii) SL with policy-only RL model, and (iii) SL with end-to-end RL model.  Table \ref{tab:eval_result_human} lists the mean and standard deviation of human evaluation scores over all system turns: end-to-end optimization with RL clearly improves the quality of the model according to human judges.

    \begin{table}[th]
    \caption{Human evaluation results with mean and standard deviation of crowd worker scores.}
    \label{tab:eval_result_human}
    \centering
    \begin{tabular}{l|c|c|c}
    \hline
    \textbf{Model} & SL            & SL + policy-only RL         & SL + end-to-end RL      \\ \hline
    \textbf{Score} & 3.987 $\pm$ 0.086 & 4.261 $\pm$ 0.089 & 4.394 $\pm$ 0.087 \\ \hline
    \end{tabular}
    \end{table}
    
    \vspace*{-0.5ex}

\section{Conclusions}
	In this work, we propose a neural network based task-oriented dialogue system that can be trained end-to-end with supervised learning and deep reinforcement learning. We first bootstrap a dialogue agent with supervised training by learning directly from task-oriented dialogue corpora, and further optimize it with deep RL during its interaction with users. We show in the experiments that deep RL optimization on top of the supervised training model leads to significant improvement on task success rate and reduction in dialogue length comparing to supervised training baseline model. The simulation and human evaluation results further illustrate benefits of performing end-to-end model training with deep RL comparing to component-wise optimization. 

\bibliography{sample}
\bibliographystyle{unsrt}

\end{document}